\documentclass{article}
\usepackage{spconf,amsmath,graphicx}
\usepackage{booktabs}

\title{W2KPE: Keyphrase Extraction with Word-Word Relation}
\name{\begin{tabular}{c}Wen Cheng$^{*}$, Shichen Dong\sthanks{Equally contributed}, Wei Wang\end{tabular}}

\address{Nanjing University, Nanjing, China}

\begin{document}
%\ninept
%
\maketitle

\begin{abstract}
This paper describes our submission to ICASSP 2023 MUG Challenge Track 4, Keyphrase Extraction, which aims to extract keyphrases most relevant to the conference theme from conference materials. We model the challenge as a single-class Named Entity Recognition task and developed techniques for better performance on the challenge:
For the data preprocessing, we encode the split keyphrases after word segmentation. In addition, we increase the amount of input information that the model can accept at one time by fusing multiple preprocessed sentences into one segment.
We replace the loss function with the multi-class focal loss to address the sparseness of keyphrases. Besides, we score each appearance of keyphrases and add an extra output layer to fit the score to rank keyphrases. 
Exhaustive evaluations are performed to find the best combination of the word segmentation tool, the pre-trained embedding model, and the corresponding hyperparameters. With these proposals, we scored 45.04 on the final test set.
\end{abstract}
\begin{keywords}
Natural Language Processing, Keyphrase Extraction
\end{keywords}

\section{Introduction}
\label{sec:intro}
The Keyphrase Extraction (KPE) track requires extracting top-$k$ key phrases from a document that reflect its main content. It is essential in many applications, e.g., for document classification. Conventional keyphrase extraction methods, which are mainly based on word frequency, are not effective on complex and long-form documents. W2NER \cite{li2022unified} provides a strategy for Name Entity Recognition (NER), it demonstrates excellent performance on this task by introducing \textit{Next-Neighboring-Word} and \textit{Tail-Head-Word-*} relationships in named entities. Based on its structure, we explore the effect of training data preprocessing and post-processing on W2NER method with the specific task requirements for KPE. The Alimeeting4MUG corpus dataset provided by MUG is texts transcribed from real-time meetings. According to the characteristics of the data, we first perform word segmentation and stop word removal and encode the split keyphrases. On this basis, we increase the amount of information that the model can receive at one time so that the model can better understand the context. In order to sort the output of keyphrases and solve the problem of imbalanced sample distribution, we designed a loss function obtained by the weighted sum of the two parts of classification and regression so that the model can be adaptively adjusted according to the samples. In addition, we also compared different pre-trained models to select the best model and its corresponding parameters.

\section{System description}
\label{sec:sysdesc}

\begin{figure}[htb]
\begin{minipage}[b]{1.0\linewidth}
  \centering
  \centerline{\includegraphics[width=8.5cm]{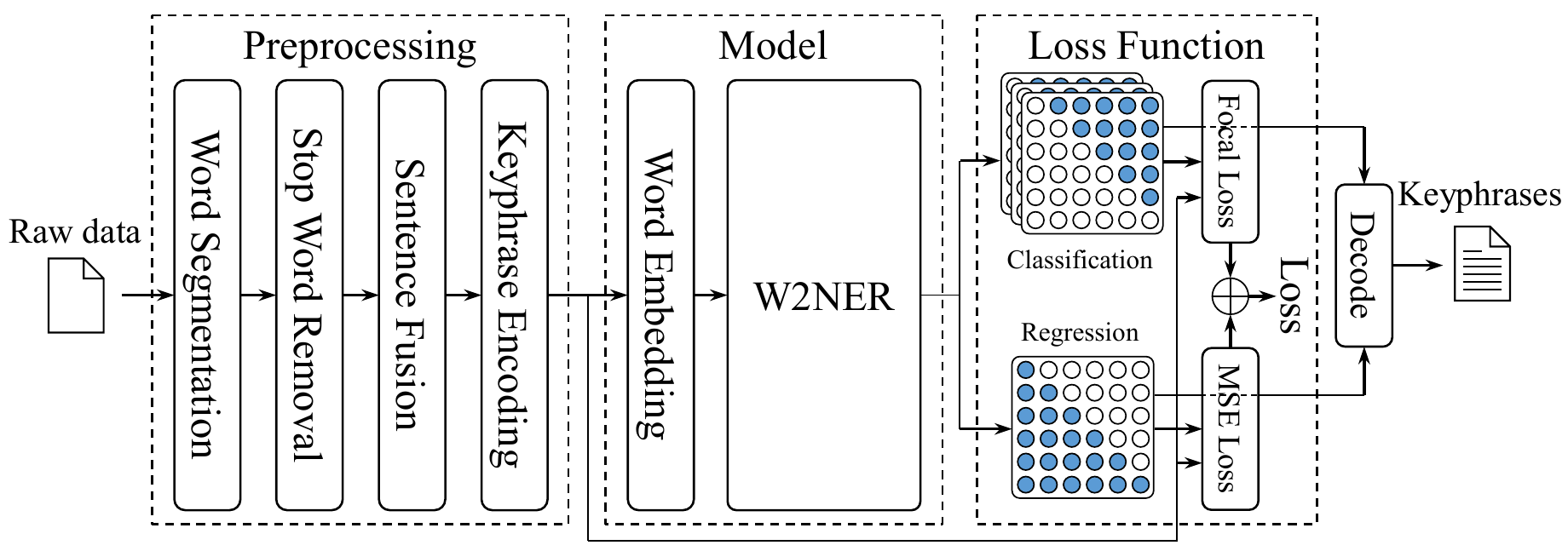}}
\end{minipage}
\caption{System architecture}
\label{fig:arch}
\end{figure}

Figure \ref{fig:arch} illustrates our overall system design, which we call W2KPE.
\subsection{Data preprocessing}
\label{sec:datapre}

The transcripts of meeting recordings in the AliMeeting4MUG dataset exhibit a high degree of oral informality, characterized by disfluencies, redundancies, omissions, and frequent deployment of modal particles. Measures should be taken to align the oral corpus with the distribution of the training dataset of the pre-trained embedding model.

\textbf{Word segmentation and stop word removal.} First, we identify and rectify all instances of stuttering by reducing consecutive occurrences of the same character that appear three or more times in a row to a single occurrence. Then, we segment the text using the open-source mandarin word segmentation tool THULAC \cite{yin2016multi}. Finally, we remove meaningless stop words to address the issue of excessive use of modal particles and to increase information density.

\textbf{Sentence fusion.} To enhance the model's comprehension of the context, it is essential to increase the amount of information fed to the model within an iteration. After the word segmentation and stop word removal, we merge multiple processed sentences into one segment while maintaining integrity at the sentence level. A threshold is needed here as a hyperparameter to determine the maximum number of words in a segment. Experiments show that the sentence fusion method significantly improves the model's performance.

\textbf{Keyphrase encoding.} We observe that keyphrases do not appear in their complete and contiguous form in many cases, but rather, they appear partially or with an intervening word. Since the W2NER model is able to recognize discontinuous named entities, we encode keyphrases that are partially presented, appear discontinuously, or are separated in the word segmentation stage in the input format of W2NER and score each appearance of keyphrases based on its completeness.

\vspace{-0.05in}
\subsection{Model design}
\vspace{-0.05in}
\label{sec:modeldesign}

We model KPE as a single-class NER task and enhance W2NER for better performance on the challenge.

\textbf{Focal loss.} To address the problem of imbalanced class distribution caused by the sparseness of keyphrases, we replace the loss function with the multi-class focal loss \cite{lin2017focal}, which focuses training on a sparse set of hard examples.

\textbf{Keyphrase scoring.} The W2NER model can only identify named entities, but not score them. To rank keyphrases, we add an extra output layer to fit the score of each appearance of keyphrases we assigned in the keyphrase encoding stage. A keyphrase's final score is determined by the sum of its scores for each appearance, other aggregation methods (e.g. averaging) are evaluated but yield lower performances.

The final loss function is the weighted sum of the focal loss and the Mean Squared Error of keyphrase scoring, which we formulate as follows, where $p_c^*$ equals $p_c$ if $c$ is the correct class, $1-p_c$ otherwise, $p_c$ is the output confidence of class $c$, $p_k$ is the output score of keyphrase $k$, $y_k$ is the ground-truth score we assigned previously, $\alpha$ and $\gamma$ are two hyperparameters which we use 0.99 and 2 respectively.

\vspace{-0.2in}
\begin{align*}
    \text{loss}&=\alpha\sum_{c\in\text{classes}}-\left(1-p_c^*\right)^\gamma\log\left(p_c^*\right) \\
               &+\left(1-\alpha\right)\sum_{k\in\text{keyphrases}}\left(p_k-y_k\right)^2 \label{eq:eq1}\tag{1}
\end{align*}

\section{Experiments and result analysis}
\label{sec:pagestyle}

\subsection{Experimental Setup}
\vspace{-0.05in}
We use the pre-trained \textit{ERNIE-3.0-base} as the word embedding model and train other parts of W2KPE from scratch, which has only 106M parameters in total. The word segmentation tool we use is THULAC, and the learning rate is set to 0.001. We trained our system with PyTorch on Tesla V100, the batch size is set to 10. The overall score is the average of exact and partial f1 scores of top-$k$ keyphrases, $k$ equals 10, 15, and 20. All ablation experiments are evaluated on dev data set, the results are shown in Table \ref{tab:training_details}. 

\subsection{Results and Analysis}
\vspace{-0.05in}
\textbf{Effect of data preprocessing.}

Compared with the experiment without sentence fusion, the W2KPE with the threshold value of 500 increases the score by $5.86$. However, it must be pointed out that the threshold cannot be increased blindly. With limited VRAM, increasing the threshold will inevitably reduce the batch size, and a too-small batch size will reduce the score instead. The keyphrase encoding also plays an important role, removing it drops the score by $2.53$.

\begin{table}
\small
\caption{Performance comparison among different setups}
\vspace{6pt}
\label{tab:training_details}
\centering
\begin{tabular}{lc}
\toprule
\multicolumn{1}{c}{\textbf{Experimental Config}} & \multicolumn{1}{c}{\textbf{Score}} \\
\midrule
\textbf{W2KPE} & \textbf{47.69}  \\
\midrule
\quad- Sentence Fusion  & 41.83(-5.86) \\
% \midrule
\quad- Keyphrase Encoding  & 45.16(-2.53)   \\
\midrule
\quad- Focal Loss & 46.94(-0.75) \\
\quad- Keyphrase Scoring & 47.05(-0.64) \\
\midrule
\ Baseline &  41.48  \\
\bottomrule
\end{tabular}
\vspace{-6pt}
\end{table}

\textbf{Effect of model design.}
Although hard cases are infrequent in the entire dataset, the use of focal loss still results in a modest improvement in the score. The fine-grained ranking of output keyphrases brought about by keyphrase scoring leads to a further elevation of the score.

The final version of W2KPE is the combination of all the best setups we mentioned. It scores $47.69$ on the dev data set, which is significantly higher than the baseline system.
\vspace{-0.05in}
\section{Conclusions}
\vspace{-0.05in}
In this paper, we proposed four notable improvements. Sentence fusion and keyphrase encoding are used in the data preprocessing stage, and we applied focal loss and keyphrase scoring for model design. These proposals have achieved significant score improvements on the final test set.

\bibliographystyle{IEEEbib}
\bibliography{strings,refs}

\end{document}